\documentclass[runningheads]{llncs}
\usepackage{hyperref}
\usepackage{algorithmic}
\usepackage{graphicx}
\usepackage{caption}
\usepackage{textcomp}
\usepackage{xcolor}
\usepackage{multirow}
\usepackage{subcaption}
\usepackage{verbatim}
\usepackage{multicol}
\usepackage{tabularx}
\usepackage{makecell}
\usepackage{booktabs}
\usepackage{amsmath}
\usepackage{rotating} 
 \usepackage[T1]{fontenc}
\usepackage{graphicx}
\begin{document}
\title{Context-aware Entity-Relation Extraction for Threat Intelligence Knowledge Graphs}

\titlerunning{CTiKG Framework}

\author{Inoussa Mouiche \and
Sherif Saad }
\author{Inoussa Mouiche\inst{1,2}\orcidID{0009-0008-8024-7631} \and
Sherif Saad\inst{1,2}\orcidID{0000-0002-5506-5261} }
%
\authorrunning{I. Mouiche and S. Saad}
%
\institute{School of Computer Science, University of Windsor, ON, Canada \and
\email{\{mouiche,shsaad\}@uwindsor.ca}}

\maketitle              
\begin{abstract}
 Cybersecurity Knowledge Graphs (CKGs) unify diverse Cyber Threat Intelligence (CTI) sources into structured, queryable formats, offering scalable solutions for automating proactive and real-time security responses. Their increasing adoption has significantly enhanced the workflow and decision-making efficiency of security professionals. However, constructing CKGs requires extracting entity–relation triples from unstructured CTI reports, a task hindered by complex report structure, domain-specific language, and semantic ambiguity. As a result, existing pipeline-based approaches often suffer from error propagation, reducing extraction accuracy and limiting generalizability.
This paper introduces the Context-aware Threat Intelligence Knowledge Graph (CTiKG) framework, a pipeline architecture designed to accurately extract and classify threat entities and their relationships from CTI reports. CTiKG incorporates hybrid NLP models that leverage SecureBERT$^+$ contextual embeddings and expert knowledge from a domain ontology to reduce misclassifications and mitigate cascading errors. Experiments on the DNRTI-AUG-STIX2 dataset, which comprises 21 entity types aligned with STIX 2.1, demonstrate significant improvements over state-of-the-art baselines, yielding 3–4\% gains in NER and up to 8\% in RE performance, based on precision, recall, and F1-score. Additional validation on DNRTI and STUCCO benchmarks confirms the framework’s robustness and practical applicability. All datasets, including the curated DNRTI-AUG-STIX2, are released on GitHub to foster reproducibility and further research.

\end{abstract}
\section{Introduction}

The rise of stealthy Advanced Persistent Threats (APTs) underscores the need for adaptive defense strategies. A recent breach disclosed by MITRE, affecting over 1,700 organizations, illustrates that no entity is immune to targeted cyberattacks~\cite{MITRE,LexCrumpton}. To anticipate such threats, security teams rely on unstructured cyber threat intelligence (CTI) reports, which capture insights on threat actors, tactics, and attack patterns. Cybersecurity Knowledge Graphs (CKGs) have emerged as a promising solution, transforming unstructured CTI into structured intelligence that supports automation, real-time analysis, and proactive defense~\cite{OpenCyKG,RelExt}. By integrating diverse sources, CKGs enhance situational awareness, improve threat understanding, and enable predictive defense.

Constructing CKGs from CTI reports requires two NLP tasks: named entity recognition (NER) and relation extraction (RE). Joint extraction (JE) models integrate both tasks but often suffer from feature confusion and overlapping relations~\cite{ZuoYali,ZexuanZhong}, while pipeline extraction (PE) allows modular optimization but is vulnerable to error propagation from entity to relation classification~\cite{CyberEntRel,CyberRel}. Both approaches struggle with the narrative complexity and language ambiguity of CTI, reducing the precision and generalizability of CKGs.

To address these issues, Mouiche and Saad~\cite{TiKG} introduced TiKG, a pipeline framework leveraging SecureBERT~\cite{SecureBERT} embeddings and a domain ontology. Despite strong performance, TiKG’s NER component (a TDD-softmax classifier) can produce inconsistent BIO tags~\cite{Lample}, and its RE model, based on a generic language model, struggles with cybersecurity-specific jargon. As a result, TiKG remains susceptible to error propagation, particularly on large and complex datasets.
This paper extends TiKG with the following contributions:
\begin{itemize} 
\item A novel NER architecture combining SecureBERT$^+$~\cite{SecureBERTPlus} embeddings with a CRF layer, achieving up to a 4\% F1 improvement across 21 entity categories.
\item A new RE architecture, base-SecureBERT$^+$, which integrates domain-specific contextual embeddings and ontology-based error control, improving F1 by up to 8\%.
\item A comprehensive evaluation on the DNRTI-AUG-STIX2 dataset, with further validation on DNRTI~\cite{DNRTI} and STUCCO~\cite{stucco}, demonstrating effectiveness and generalizability.
\item The release of three benchmark datasets DNRTI-AUG-STIX2, DNRTI, and STUCCO on GitHub\footnote{https://github.com/imouiche/Threat-Intelligence-Knowledge-Graphs} to support reproducibility and future research.
\end{itemize}

Paper structure: Section 2 reviews related work, Section 3 presents the proposed framework and experiments, Section 4 discusses CKG construction, and Section 5 concludes with key findings and future directions.

\section{Related Work}
CTI-to-CKG extraction methods typically follow either pipeline (PE) or joint extraction (JE) paradigms.

\subsection{Joint Extraction}
JE unifies NER and RE in a single model to reduce error propagation and improve accuracy. Many approaches use multi-task learning with shared encoders and transformer-based architectures. Examples include ERBTF~\cite{WangXiaodi}, which combines relation and word embeddings; CyberRel~\cite{YongyanGuo,CyberRel} and CyberEntRel~\cite{CyberEntRel}, which model JE as sequence labeling with BiGRU/CRF layers; and Liu et al.~\cite{CTI-JE}, who framed JE as a table-filling task with SecBERT. More recent models include CTI-TFN~\cite{CTI-TFN}, a Fourier-based joint model, ITIRel~\cite{ITIRel} for overlapping IoT relations, and TIJERE~\cite{TIJERE}, a data-centric JE framework.  

JE models streamline extraction but lack flexibility; modifying one task often requires retraining the entire system. They also struggle with overlapping entities~\cite{ZuoYali}, feature confusion~\cite{PipeVsJoint}, and limited domain adaptation when relying on generic LMs~\cite{TiKG}.  

This work instead extends the PE approach of TiKG~\cite{TiKG}, decoupling NER and RE for modular design. Each subtask leverages SecureBERT embeddings, with a domain ontology guiding entity–relation mapping to improve classification and CKG quality.

\subsection{Pipeline Extraction}
In PE, NER identifies entities, and RE generates relation triples. Studies adopting this paradigm include Gasmi et al.~\cite{HoussemGasmi} (BiLSTM-CRF), HINT~\cite{HINT} (attention BiLSTM-CRF + heterogeneous graphs), Vulcan~\cite{Vulcan} (BERT-BiLSTM-CRF for ransomware), and STIXnet~\cite{STIXnet}, which integrates regex, dependency parsing, and neural RE. Other methods include SVM-based vulnerability extraction~\cite{VarishM}, entity–coreference integration (EEMAP~\cite{YongfeiLi}), and OIE-based approaches like Open-CyKG~\cite{OpenCyKG}.  

Piplai et al.~\cite{AritranPiplai} structured a CKG using UCO 2.0, while our framework enhances TiKG~\cite{TiKG} with SecureBERT$^+$-based NER/RE. These upgrades improve domain relevance, mitigate misclassification, and reduce error propagation. Evaluations on DNRTI-AUG-STIX2 and validation on DNRTI and STUCCO datasets show 3–4\% higher NER F1 and 8\% higher RE F1, establishing CTiKG as a robust, security-aware framework for threat analysis, profiling, and defense strategy development.

\section{CTiKG: Context-aware Threat Intelligence Knowledge Graph Framework}
\begin{figure}[htbp]
\centerline{\includegraphics[width=\textwidth, height=2.7in]{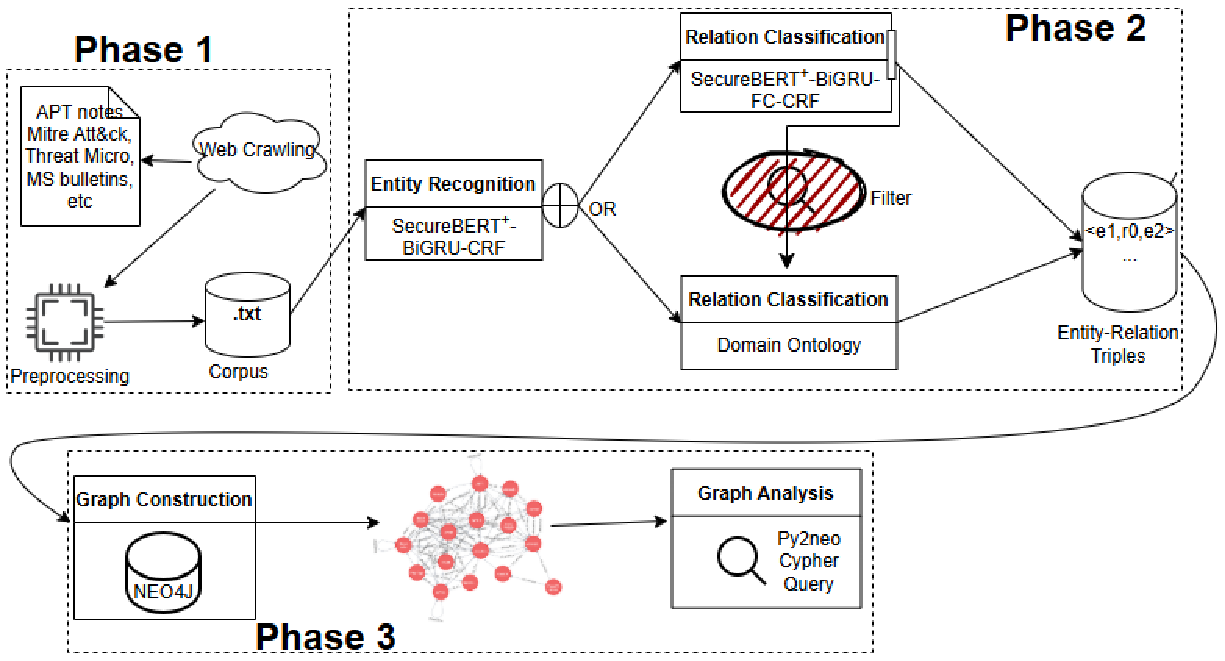}}
\caption{Context-aware Threat Intelligence Knowledge Graph Framework: CTiKG.}
\label{fig:CTiKG}
\end{figure}

This section presents the proposed CTiKG framework, a novel CTI extraction pipeline for constructing CKGs, inspired by Mouiche and Saad~\cite{TiKG}. As shown in Figure~\ref{fig:CTiKG}, the framework comprises three main phases:
\label{sec:CTiKGFramewrok}
\begin{itemize}
\item \textbf{Phase 1: Data Collection and Processing}: Reports are crawled from CTI repositories (e.g., APT repository, MITRE ATT\&CK, TrendMicro) and converted into plain-text using HTMLParser, PDFLib, and pdftotext. Preprocessing sanitizes text (e.g., obfuscating URLs and emails) to produce .txt files simulating raw inputs for the extraction pipeline.
\item \textbf{Phase 2: Entity–Relation Extraction}:  
This phase extracts threat entities and their relationships from the processed text through two modules: \textit{Entity Recognition} and \textit{Relation Classification}. The former identifies and labels entities in the Phase 1 reports according to predefined categories, while the latter uses either a rule-based \textit{Domain Ontology} or a hybrid method that combines SecureBERT$^+$-BiGRU-CRF with the ontology. For small to medium datasets, the ontology alone can generate reliable \textit{Entity–Relation Triples}. For larger datasets, the hybrid method improves scalability and accuracy: the ML model predicts relations, and misclassified instances are corrected through the ontology.

\item \textbf{Phase 3: Construction and Analysis of CTiKG}:
This final phase builds and evaluates the knowledge graph using the entity-relation triples from Phase 2. Neo4j stores entities as nodes and relations as edges. The graph is populated and analyzed using the Neo4j graph data science library, which also enables query-based retrieval to assess the quality and effectiveness of the constructed CTiKG.
\end{itemize}
The proposed context-aware pipeline enhances the CTI workflow by transforming unstructured reports into a structured, queryable format that reveals hidden patterns and supports timely, informed decision-making for security stakeholders. The following sections detail the core components of the CTiKG framework and highlight the key enhancements introduced over the original TiKG model~\cite{TiKG}.

\subsection{Entity Recognition}
In pipeline extraction, errors in NER directly propagate to RE, reducing the quality of extracted triples and the resulting CKG. Ensuring high NER accuracy is therefore critical. In the TiKG framework~\cite{TiKG}, SecureBERT-BiLSTM-TDD was adopted for entity recognition: SecureBERT provided contextual embeddings, BiLSTM captured token dependencies, and the TDD layer performed token-level classification. However, since TDD predicts tokens independently with softmax, it often produces invalid spans (e.g., missing B-tags or misaligned I-sequences)~\cite{Lample}, which introduce noise into RE and degrade relation classification. To address this limitation, we replace TDD with a CRF layer to enforce valid tag transitions and improve sequence-level consistency~\cite{CRF}. We also integrate SecureBERT$^+$~\cite{SecureBERTPlus}, a domain-adapted variant that improves contextual representation, and combine BiLSTM and BiGRU sequential modeling to further enhance dependency capture and robustness in CTI reports.
\begin{figure}[htbp]
\centerline{\includegraphics[width=.65\textwidth, height=2.8in]{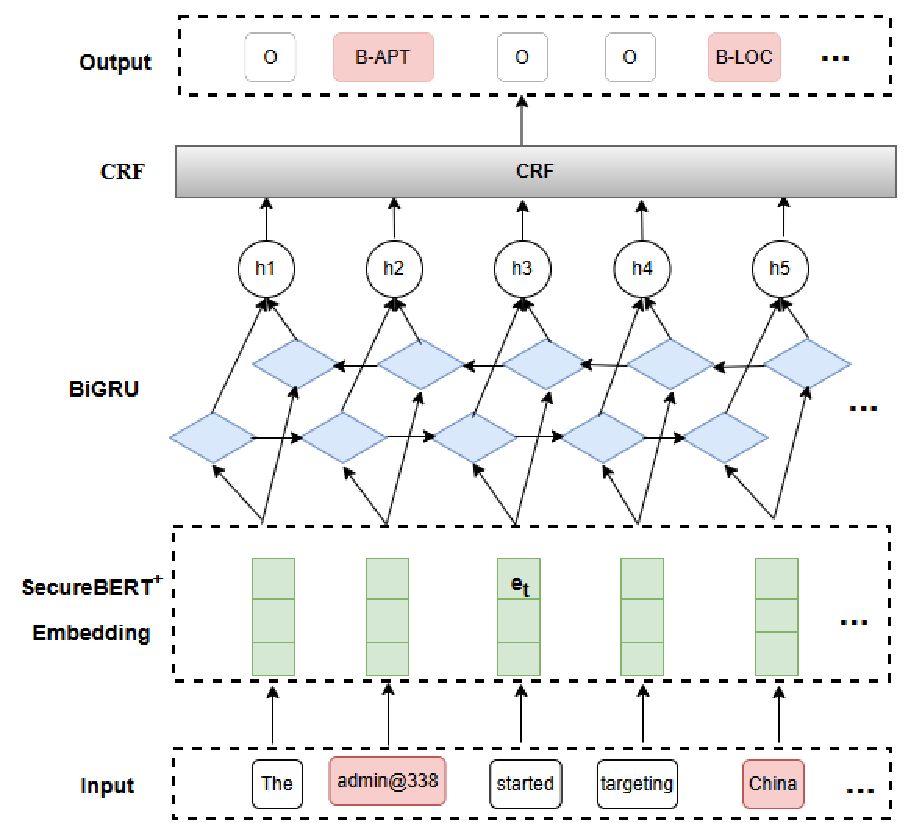}}
\caption{NER model's archicteure (SecureBERT$^+$-BiGRU-CRF).}
\label{fig:secureBERT-BiGRU-CRF}
\end{figure}

Figure \ref{fig:secureBERT-BiGRU-CRF} presents our hybrid model, which can be described as follows:
\begin{itemize}
    \item \textbf{Input to SecureBERT$^+$ Embedding:}
    \begin{equation}
        \mathbf{e}_t = \text{SecureBERT$^+$}(\mathbf{x}_t) \quad \forall t \in \{1, \ldots, T\},
    \end{equation}
    Where \(\mathbf{x}\) is the input sequence of length $T$, \(\mathbf{x}_t\) represents the $t$th input tokens, and \(\mathbf{e}_t\) is the corresponding embedding produced by SecureBERT$^+$.
    \item \textbf{SecureBERT$^+$ Embedding to BiGRU:}
    \begin{equation}
        \mathbf{h}_t = \text{BiGRU}(\mathbf{e}_t)
    \end{equation}
Here, \(\mathbf{h}_t\) is the hidden state at time step \(t\) produced by the bidirectional GRU.
\item \textbf{BiGRU to CRF Emissions:}
\begin{equation}
    \mathbf{o}_t = \mathbf{W} \mathbf{h}_t + \mathbf{b}
\end{equation}
Where \(\mathbf{W}\) is the weight matrix and \(\mathbf{b}\) is the bias vector. \(\mathbf{o}_t\) represents the emissions for the CRF layer.
\item \textbf{CRF Layer:}
\begin{equation}
    \mathbf{y} = \arg\max_{\mathbf{y}'} P(\mathbf{y}' | \mathbf{o})
\end{equation}
Where \(P(\mathbf{y}' | \mathbf{o})\) is the conditional probability of the label sequence \(\mathbf{y}'\) given the emissions \(\mathbf{o}\). 
In a CRF, the probability of a particular label sequence \(\mathbf{y}\)
given the emissions \(\mathbf{o}\) is given by:
\begin{equation}
    P(\mathbf{y} | \mathbf{o}) = \frac{\exp(\text{z}(\mathbf{o}, \mathbf{y}))}{\sum_{\mathbf{y}'} \exp(\text{z}(\mathbf{o}, \mathbf{y}'))}
\end{equation}

where $\text{z}(\mathbf{o}, \mathbf{y})$ is given by:

\begin{equation}
    \text{z}(\mathbf{o}, \mathbf{y}) = \sum_{t=1}^{T} \left( \mathbf{A}_{y_{t-1}, y_t} + \mathbf{o}_{t, y_t} \right)
\end{equation}
Here, \( T \) is the length of the input sequence \(\mathbf{x}\), \(\mathbf{A}\) is the transition matrix, \(\mathbf{A}_{y_{t-1}, y_t}\) represents the score of transitioning from label \(y_{t-1}\) to label \(y_t\), and \(\mathbf{o}_{t, y_t}\) is the emission score for label \(y_t\) at time step \(t\). For the CRF, the loss function \(\mathcal{L}\) is typically the negative log-likelihood given by:

\begin{equation}
   \mathcal{L} = -\log P(\mathbf{y} | \mathbf{o}) = -\log \left( \frac{\exp(\text{z}(\mathbf{o}, \mathbf{y}))}{\sum_{\mathbf{y}'} \exp(\text{z}(\mathbf{o}, \mathbf{y}'))} \right) \label{eq:log7}
\end{equation}
Where,
\begin{equation}
{\sum_{\mathbf{y}'} \exp(\text{z}(\mathbf{o}, \mathbf{y}'))} = \sum_{\mathbf{y}'} \exp\left(\sum_{t=1}^{T} \left( \mathbf{A}_{y_{t-1}', y_t'} + \mathbf{o}_{t, y_t'} \right)\right) \label{eq:log}
\end{equation}
The Eq. \ref{eq:log} is also called the partition function and normalizes the probability distribution over all possible label sequences \(\mathbf{y}'\).
\end{itemize}

With the mathematical formulation established, we proceed to implement and evaluate the SecureBERT$^+$-BiGRU-CRF model, comparing its performance with existing extraction models in the literature.

\subsection{NER Implementation}
\subsubsection{\textbf{Datasets}}  
This study uses DNRTI-AUG-STIX2, an augmented version of the DNRTI-STIX2 threat intelligence NER dataset recently released by~\cite{TI-NERmerger}. DNRTI-STIX2 was designed to align with the STIX 2.1 standard~\cite{STIX}, covering a diverse range of STIX domain and observable objects and making it well-suited for entity–relation extraction. Unlike earlier datasets with narrow coverage, it includes 21 entity categories distributed across 6,580 sentences (see Table~\ref{tab:dataset_info}). However, it suffers from severe class imbalance, with some categories having fewer than five instances. Prior work~\cite{Pedregosa} suggests that at least 50 samples per class are needed for models to learn contextual features effectively. To mitigate this limitation, we applied targeted data augmentation to transform DNRTI-STIX2 into DNRTI-AUG-STIX2.

\begin{itemize}
    \item \textbf{Data Augmentation (DA):} To increase representation of under-sampled classes, we annotated additional APT report sentences curated for CTiKG testing. A total of 1,367 sentences were manually labeled by three students (two master's and one PhD, all with security backgrounds) using the same 21 entity types and BIO tagging scheme as DNRTI-STIX2. Sentences were selected to ensure coverage of less frequent entity classes. Annotation guidelines were established through two calibration meetings, after which the dataset was divided (50\% PhD, 25\% each master's student). Conflicts were resolved by majority vote, with the PhD student’s vote weighted at 50\%. The resulting annotations were added to DNRTI-STIX2.
    
    \item \textbf{Data Consolidation (DC):} Rare classes were merged into broader categories. For example, SHA1, SHA2, and MD5 were consolidated into a single entity type, \texttt{HASH}, representing hash algorithms.
\end{itemize}

\begin{table}[h]
\centering
\caption{DNRTI-STIX2 transformed into DNRTI-AUG-STIX2 using Data Augmentation (DA) and Data Consolidation (DC).}
\begin{tabular}{|p{3cm}|p{2cm}|p{3cm}|p{1cm}|}
    \toprule
    \multicolumn{2}{|c|}{\textbf{DNRTI-STIX2}} & \multicolumn{2}{|c|}{\textbf{DNRTI-AUG-STIX2}} \\
    \hline
    Entity Type & Count & Entity Type & Count\\
    \midrule
    ACT & 8871 & ACT & 9070\\
    APT & 5867 & APT & 5906\\
    DOM & 41 & DOM & 435\\
    EMAIL & 32 & EMAIL & 66\\
    ENCR & 41 & ENCR & 242\\
    FILE & 2105 & FILE & 2458\\
    IDTY & 5573 & HASH & 483\\
    IP & 14 & IDTY & 5845\\
    LOC & 3520 & IP & 229\\
    MAL & 3294 & LOC & 3615\\
    MD5 & 5 & MAL & 3924\\
    OS & 242 & OS & 600\\
    PROT & 160 & PROT & 519\\
    SECTEAM & 1953 & SECTEAM & 2026\\
    SHA1 & 2 & TIME & 3039\\
    SHA2 & 6 & TOOL & 3898 \\
    TIME & 2675 & URL & 100\\
    TOOL & 3062 & VULID & 803\\
    URL & 6 & VULNAME & 1312\\
    VULID & 747 & & \\
    VULNAME & 1243 & &\\
    \bottomrule \\
    \# of sentences & 6580 & \# of sentences & \textbf{7947}\\
    \hline
     vocab\_size & 9444 & vocab\_size & \textbf{11741}\\
    \hline
    \# of entity types & 21 & \# of entity types & \textbf{19}\\
    \bottomrule
\end{tabular}
\label{tab:dataset_info}
\end{table}
The results after applying DA and DC are shown in the last two columns of Table \ref{tab:dataset_info}. This increased 1367 sentences and 2297 new vocabularies. Additionally, the number of entities was reduced from 21 to 19 due to merging SHA1, SHA2, and MD5 into a single HASH entity type. The resulting augmented dataset as DNRTI-AUG-STIX2 will be made available on our GitHub to support research in the field.\\
To scale the generalization and reproducibility, we validate our models on two additional open-source NER datasets: DNRTI~\cite{DNRTI} and STUCCO~\cite{stucco}.
\begin{itemize}
\item DNRTI: This dataset contains 175,220 tokens in 6,592 sentences describing APT reports. It features 13 entity categories: HackOrg, OffAct, SamFile, SecTeam, Tool, Time, Purp, Area, Idus, Org, Way, Exp, and Features.
\item STUCCO: This dataset contains CVE and NVD descriptions, totaling 680,764 tokens represented in 15,192 sentences. It encompasses 15 entity types: application, cveID, edition, file, function, hardware, method, OS, parameter, programming language, relevant term, update, vendor, and version.
\end{itemize}

\subsubsection{\textbf{Training and Evaluation}}
For training and evaluation, we divided the dataset into training, validation, and testing sets with a split ratio of 70\%, 15\%, and 15\% respectively. Table \ref{tab:modelparams} provides the base model parameter settings, as reported in~\cite{TiKG}. It is important to note that we used the same parameters for the RoBERTa base model as those used with SecureBERT variants. 
\begin{table}[h]
\centering
\caption{NER Models' parameter settings.}
\begin{tabular}{|l|c|c|c|}
\hline
\textbf{parameters} & \textbf{SecureBERT} & \textbf{BERT} & \textbf{BiLSTM}\\ \hline
batch size    & 8  & 8 & 16\\ \hline
dropout  & 0.2 & 0.2 & 0.2 \\ \hline
epsilon & 1e-8& 1e-8 & - \\ \hline
initial learning rate & 5e-5& 5e-5 & 5e-5\\ \hline
hidden layer size & 128 $\times$ 2 & 128 $\times$ 2 & 100 $\times$ 2 \\ \hline
embedding size & 768 & 768 & 300\\ \hline
number of epochs & 4 & 4 & 10\\ \hline
optimizer & AdamW & AdamW & AdamW\\\hline
\end{tabular}
\label{tab:modelparams}
\end{table}
To demonstrate the performance of our proposed NER models, we implemented other state-of-the-art approaches from scratch to serve as baselines. We employed standard metrics such as Precision (P), Recall (R), and F1-score (F1) for evaluation and comparison.
\begin{table*}[h]
\centering
\caption{NER models: Evaluation results, comparisons and validation across DNRTI-AUG-STIX2, DNRTI, and STUCCO datasets.}
\begin{tabular}{|p{4.2cm}|p{.8cm}|p{.8cm}|p{.5cm}|p{.6cm}|p{.6cm}|p{.7cm}|p{.65cm}|p{.65cm}|p{.65cm}|}
\hline
 & \multicolumn{3}{c}{\textbf{DNRTI-AUG-STIX2}} & \multicolumn{3}{|c|}{\textbf{DNRTI}} & \multicolumn{3}{|c|}{\textbf{STUCCO}} \\
    \cline{2-4}  \cline{5-7} \cline{8-10}

\textbf{Models} & \textbf{P} &\textbf{R} &\textbf{F1} & \textbf{P} &\textbf{R} &\textbf{F1}& \textbf{P} &\textbf{R} &\textbf{F1} \\ \hline
BiLSTM-CRF & 0.68 & 0.70 & 0.70 & 0.67 & 0.72 & 0.71 & 0.76 & 0.77 & 0.75 \\ \hline
RoBERTa & 0.80 & 0.83 & 0.81 & 0.81 & 0.85 & 0.83 & 0.93 & 0.94 & 0.93 \\ \hline
BERT-BiLSTM-CRF~\cite{Vulcan,ZuoYali} & 0.80& 0.85 & 0.83 & 0.82 & 0.84 & 0.83 & 0.95 & 0.96 & 0.95\\ \hline
BERT-BiGRU-CRF~\cite{CyberRel}  & 0.84 & 0.87 & 0.86 & 0.85 & 0.85 & 0.85 & 0.96 & 0.96 & 0.95\\ \hline
RoBERTa-BiLSTM-TDD+Att~\cite{TiKG} & 0.87 & 0.89 & 0.88 & 0.86 & 0.88 & 0.87 & 0.96 & 0.97 & 0.96 \\ \hline
RoBERTa-BiGRU-CRF~\cite{CyberEntRel,OpenCyKG} & 0.88 & 0.89 & 0.89 & 0.89 & 0.90 & 0.89 & 0.97 & 0.97 & 0.96\\ \hline
SecureBERT-BiLSTM-TDD+Att~\cite{TiKG} & 0.89 & 0.90 & 0.90 & 0.90 & 0.91 & 0.90 & 0.97 & 0.98 & 0.97\\ \hline
SecureBERT-BiLSTM-CRF & 0.91 & 0.91 & 0.91 & 0.92 & 0.91 & 0.91 & 0.98 & 0.98 & 0.98\\ \hline
SecureBERT-BiGRU-CRF  & \textbf{0.93} & \textbf{0.92} & \textbf{0.92} & \textbf{0.93} & \textbf{0.94} & \textbf{0.93} & \textbf{0.99} & \textbf{0.99} & \textbf{0.98}\\
\hline
SecureBERT$^+$-BiGRU-CRF   & \textbf{0.93} & \textbf{0.94} & \textbf{0.93} & \textbf{0.94} & \textbf{0.95} & \textbf{0.94} & \textbf{0.99} & \textbf{0.99} & \textbf{0.98}\\
\hline
\end{tabular}
\label{tab:nerEval}
\end{table*}

The performance comparison in Table~\ref{tab:nerEval} shows that the SecureBERT$^+$-BiGRU-CRF model consistently outperforms all other NER models across the DNRTI-AUG-STIX2, DNRTI, and STUCCO datasets. It achieves the highest scores on all three datasets, with Precision (0.93), Recall (0.94), and F1-score (0.93) on DNRTI-AUG-STIX2, Precision (0.94), Recall (0.95), and F1-score (0.94) on DNRTI, and Precision (0.99), Recall (0.99), and F1-score (0.98) on STUCCO. These results confirm the superior performance and generalization capability of our proposed model in extracting cybersecurity-specific entities.
The results also demonstrate that integrating a CRF layer over SecureBERT$^+$ embeddings yields consistent improvements over earlier architectures such as SecureBERT-BiLSTM-TDD~\cite{TiKG}, achieving 3–4\% higher precision, recall, and F1-scores. This enhancement can be attributed to the CRF’s ability to model sequence-level dependencies and produce more coherent BIO tag sequences. Moreover, we observe that there is no significant performance gap between SecureBERT and its improved variant SecureBERT$^+$ in the context of entity extraction. Despite SecureBERT$^+$ achieving a 9\% improvement in masked language modeling~\cite{SecureBERTPlus}, this gain does not directly translate into enhanced NER performance, suggesting that improvements at the pretraining level may not always yield proportional benefits for downstream tasks like NER.\\
All implemented models perform strongly on the STUCCO dataset. This is not only due to better class distribution but also because STUCCO’s vulnerability descriptions exhibit consistent linguistic patterns, reducing ambiguity across entity types and improving model prediction accuracy.

\subsection{Relation Extraction Model}
The prior TiKG framework~\cite{TiKG} relied on general-purpose language models pretrained on corpora such as Wikipedia and BooksCorpus. These models fail to capture cybersecurity-specific terminology and contextual nuances in APT reports. For example, terms like ``APT28'' or ``Mimikatz'' carry precise meanings only within the cybersecurity domain, yet generic models often misclassify their roles, leading to errors that degrade knowledge graph quality. To overcome this limitation, we adopt the same security-aware backbone used in our NER model, SecureBERT$^+$-BiGRU, ensuring consistency and domain-awareness across the pipeline. Unlike NER, relation types are independent and lack sequential dependencies, so we replace the CRF layer with a TDD layer, simplifying the architecture and enabling faster inference. This design significantly improves relation classification, yielding up to a 7\% increase in F1 score compared to prior approaches.The resulting RE model architecture is illustrated in Figure~\ref{fig:secureBERT-BiGRU-FC-CRF} and formally defined as follows:
\begin{figure}[htbp]
\centerline{\includegraphics[width=.6\textwidth, height=3.in]{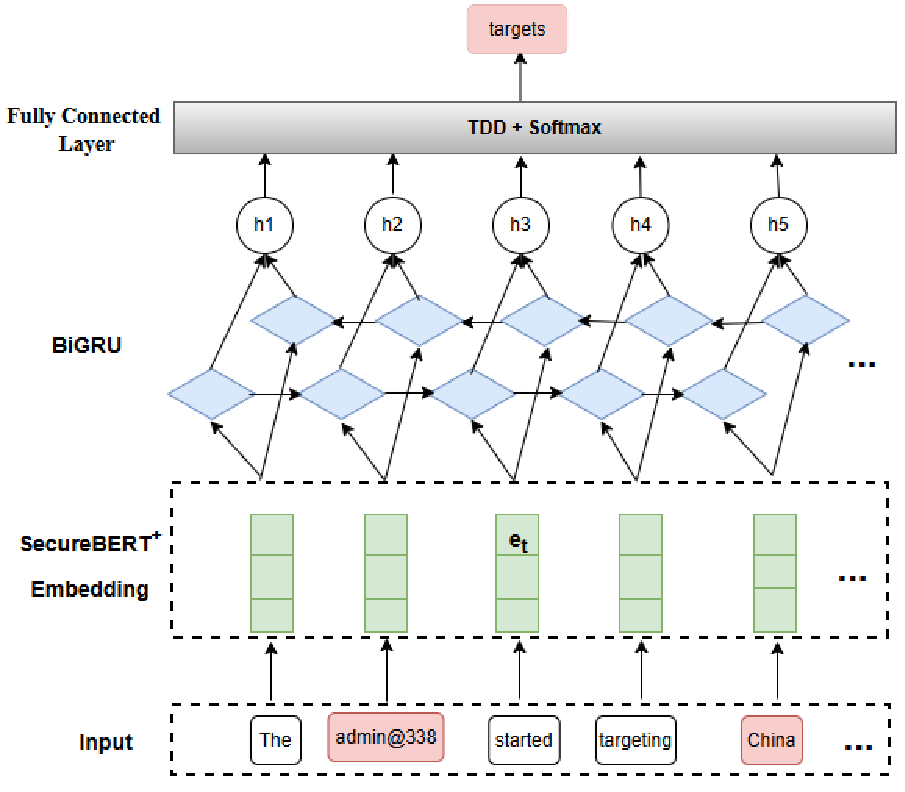}}
\caption{SecureBERT$^+$-BiGRU-TDD relation extraction model.}
\label{fig:secureBERT-BiGRU-FC-CRF}
\end{figure}
\begin{itemize}
    \item \textbf{Input to SecureBERT$^+$ Embedding:}
    \begin{equation}
        \mathbf{e}_t = \text{SecureBERT}^+(\mathbf{x}_t), \quad \forall t \in \{1, \ldots, T\},
    \end{equation}
    where \(\mathbf{x} = (x_1, \ldots, x_T)\) is the input token sequence, and \(\mathbf{e}_t\) is the contextual embedding of token \(x_t\) produced by SecureBERT$^+$.

    \item \textbf{SecureBERT$^+$ Embedding to BiGRU:}
    \begin{equation}
        \mathbf{h}_t = \text{BiGRU}(\mathbf{e}_t),
    \end{equation}
    where \(\mathbf{h}_t\) is the hidden state at time step \(t\) produced by the bidirectional GRU.

    \item \textbf{BiGRU to Fully Connected Layer (TDD):}
    \begin{equation}
        \mathbf{f}_t = \mathbf{W}_{\text{fc}} \mathbf{h}_t + \mathbf{b}_{\text{fc}},
    \end{equation}
    where \(\mathbf{W}_{\text{fc}}\) and \(\mathbf{b}_{\text{fc}}\) are the parameters of the time-distributed dense layer, and \(\mathbf{f}_t\) is the logit vector at time step \(t\).

    \item \textbf{Softmax Output for Relation Classification:}
    \begin{equation}
        \hat{y}_t = \text{softmax}(\mathbf{f}_t),
    \end{equation}
    where \(\hat{y}_t\) is the predicted probability distribution over relation labels for the entity pair at time step \(t\).
\end{itemize}

Training and evaluating the relation extraction model requires an annotated dataset with relation labels. To support this, we introduce a domain ontology specifically designed for the DNRTI-AUG-STIX2 dataset. This ontology aids both in labeling the dataset and in post-processing model outputs, improving relation classification accuracy.

\subsubsection{DNRTI-AUG-STIX2 Domain Ontology}
\begin{figure}[!h]
\centerline{\includegraphics[scale=0.8]{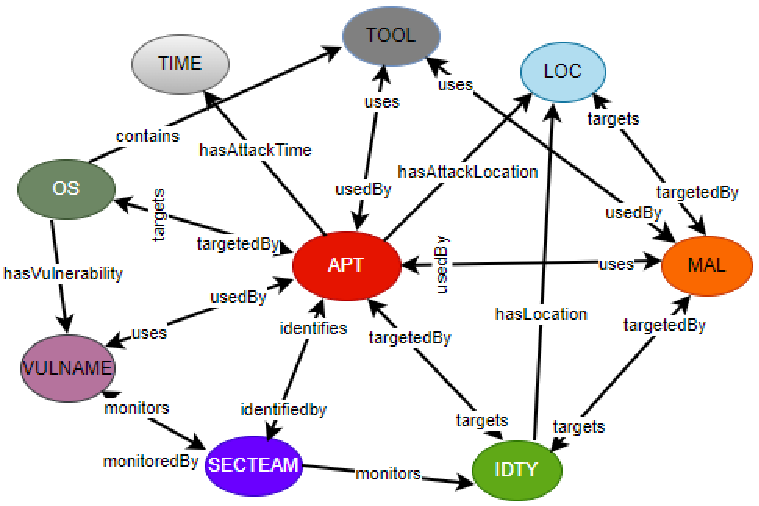}}
\caption{A sample ontology schema containing entities (nodes) and 12 relation categories (edges).}
\label{fig:sample_re_schema}
\end{figure}

Syed et al.~\cite{UCO} introduced the unified cyber ontology (UCO) 2.0 framework to standardize threat information representation. Inspired by this, we defined a domain-specific ontology tailored to the DNRTI-AUG-STIX2 dataset, capturing threat entities and their relationships. The ontology defines 15 relations to describe connections between entity pairs, improving the resulting CKG from the CTiKG framework:
\begin{itemize}
    \item affiliatedWith: APT $\rightarrow$ APT
    \item associatedWith: (HASH, VULNAME, VULID) $\rightarrow$ (EMAIL, ACT, ENCR, DOM, URL, TOOL, OS, PROT)
    \item contains: (FILE, EMAIL) $\rightarrow$ (MAL, IP, URL)
    \item hasAttackLocation: (APT, MAL, ACT) $\rightarrow$ LOC
    \item hasAttackTime: (APT, MAL, ACT) $\rightarrow$ TIME
    \item hasLocation: (IDTY, SECTEAM) $\rightarrow$ LOC
    \item hasVulnerability: (IDTY, OS, URL, DOM, PROT, FILE) $\rightarrow$ (VULID, VULNAME)
    \item identifies: SECTEAM $\rightarrow$ (APT, MAL, VULNAME, VULID, ACT)
    \item identifiedBy: (APT, MAL, VULNAME, VULID, ACT) $\rightarrow$ SECTEAM
    \item monitors: SECTEAM $\rightarrow$ (IDTY, DOM, FILE)
    \item monitoredBy: (IDTY, LOC, FILE) $\rightarrow$ SECTEAM
    \item targets: (APT, MAL, ACT) $\rightarrow$ (IDTY, DOM, VULNAME, VULID, OS) 
    \item targetedBy: (IDTY, DOM, VULNAME, VULID, OS, LOC) $\rightarrow$ (APT, MAL, ACT)
    \item uses: (APT, MAL, ACT) $\rightarrow$ (EMAIL, IP, URL, FILE, TOOL, HASH, ENCR, MAL, ACT)
    \item usedBy: (EMAIL, IP, URL, FILE, TOOL, HASH, ENCR, MAL, ACT) $\rightarrow$ (APT, MAL, ACT)
\end{itemize}
A sample schema is shown in Figure \ref{fig:sample_re_schema}, including 9 of 19 entity types and 12 of 16 relation types. We used this ontology to automatically assign relation types to entity pairs, discarding invalid ones. This enabled consistent, ontology-guided annotation of DNRTI-AUG-STIX2.

\subsubsection{Training and Evaluation}
The DNRTI-AUG-STIX2 dataset was split into training, validation, and test sets (70:15:15). Default hyperparameters for each RE model are listed in Table~\ref{tab:re_modelparams}, following~\cite{TiKG}. We reimplemented several state-of-the-art RE approaches as baselines to benchmark our models.

\begin{table}[!h]
\centering
\caption{RE Models' hyper-parameter settings.}
\begin{tabular}{|l|c|c|c|c|}
\hline
\textbf{parameters} & \textbf{SecureBERT} & \textbf{BERT} & \textbf{BiLSTM} & \textbf{CNN}\\ \hline
batch size    & 16  & 16 & 16 & 16\\ \hline
dropout  & 0.3 & 0.3 & 0.3 & 0.3\\ \hline
epsilon & 1e-8& 1e-8 & 1e-8 & 1e-8 \\ \hline
initial learning rate & 5e-5& 5e-5 & 5e-5 & 5e-5\\ \hline
hidden layer size & 100 $\times$ 2 & 100 $\times$ 2 & 100 $\times$ 2 & -\\ \hline
embedding size & 768 & 768 & 300 & 300\\ \hline
number of epochs & 4 & 4 & 50 & 100\\ \hline
optimizer & AdamW & AdamW & AdamW & AdamW\\\hline
window & - & - & - & 3 \\ \hline
pos\_dim, pos\_dist & - & - & - & 5, 50 \\ \hline
\end{tabular}
\label{tab:re_modelparams}
\end{table}
\begin{table}[!h]
\centering
\caption{RE models: Evaluation results and comparisons}
\begin{tabular}{|p{5cm}|p{1.5cm}|p{1.5cm}|p{1.5cm}|}
\hline
\textbf{Models} & \textbf{Precision} &\textbf{Recall} &\textbf{F1}  \\ \hline
Glove-CNN-TDD+Att~\cite{TiKG}   & 0.84 & 0.84 & 0.84 \\ \hline
BERT-BiLSTM-TDD+Att~\cite{TiKG}   & 0.90 & 0.89 & 0.89 \\ \hline
Glove-BiLSTM-TDD+Att~\cite{TiKG}  & 0.91 & 0.91 & 0.91 \\ \hline
RobERTa-BiGRU-CRF   & 0.94 & 0.94 & 0.94 \\ \hline
RobERTa-BiGRU-TDD   & 0.95 & 0.94 & 0.94 \\ \hline
SecureBERT-BiGRU-CRF   & \textbf{0.97} & \textbf{0.97} & \textbf{0.97} \\\hline
SecureBERT$^+$-BiGRU-CRF   & \textbf{0.98} & \textbf{0.97} & \textbf{0.97} \\\hline
SecureBERT$^+$-BiGRU-TDD   & \textbf{0.98} & \textbf{0.98} & \textbf{0.98} \\
\hline
\end{tabular}
\label{tab:reEval}
\end{table}
Table~\ref{tab:reEval} shows that SecureBERT$^+$-BiGRU-TDD achieved the best performance (Precision/Recall/F1 = 0.98), closely matched by SecureBERT-BiGRU-TDD. The added 9\% MLM gain of SecureBERT$^+$ provided no meaningful RE improvement. Similarly, CRF-based variants offered no advantage over TDD-only models, indicating that CRF layers add complexity without measurable benefit. Compared with RobERTa-BiGRU-CRF (F1 = 0.94) and Glove-BiLSTM-TDD+Att (F1 = 0.91), SecureBERT-based models show clear improvements in both precision and consistency. Overall, they outperform TiKG’s RE module~\cite{TiKG} by 6–8\% across all metrics, validating the benefit of domain-specific contextual embeddings and streamlined architectures.

\begin{figure}[htbp]
\centerline{\includegraphics[width=.7\textwidth, height=3in]{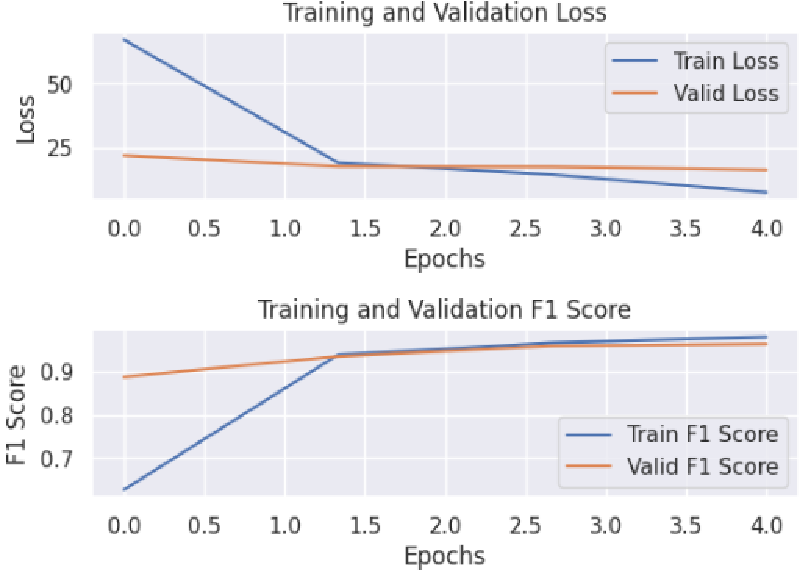}}
\caption{SecureBERT$^+$-BiGRU-TDD: training and validation losses and F1 Scores.}
\label{fig:re_train_eval_loss}
\end{figure}
Figure~\ref{fig:re_train_eval_loss} illustrates SecureBERT$^+$-BiGRU-TDD training dynamics, showing rapid loss reduction and stable validation trends, with F1 convergence near 0.96. This demonstrates effective learning and minimal overfitting. 
The early-stage divergence, characterized by a rapid decline in training loss but only a modest decrease in validation loss, indicates rapid model fitting with limited immediate generalization. However, both training and validation F1 scores continue to improve and converge around 0.96, confirming the model's ability to maintain high validation precision.
Integrating these models into the TiKG pipeline yields CTiKG, a fully security-aware system with improved NER and RE, reducing error propagation and enhancing the reliability of the constructed CKG.

The knowledge graph construction and downstream analysis based on the proposed CTiKG framework follow a similar process to its predecessor, TiKG\cite{TiKG}. These aspects are left for future work, where a comparative study of both pipelines will assess their effectiveness in supporting threat prediction and countermeasure planning, ultimately aiding security analysts in timely and informed decision-making.

\section{Conclusion}
We presented CTiKG, a context-aware threat intelligence knowledge graph framework combining SecureBERT$^+$ embeddings with BiGRU and CRF architectures for accurate NER and RE. A domain ontology further guided entity-pair classification, ensuring coherent graph construction.  
Trained on DNRTI-AUG-STIX2, our models achieved up to 3–4\% NER and 8\% RE gains over prior work, including TiKG, with validation and robustness confirmed across DNRTI and STUCCO datasets.  
Future work will extend CTiKG to heterogeneous graphs integrating multiple CTI sources, expand datasets with recent reports, and design protocols to assess graph quality under realistic threat scenarios, advancing AI-driven threat intelligence for security analysts.

\subsubsection*{Disclaimer} 
The camera-ready version of this paper was partially edited using AI-assisted tools to meet conference page limits and formatting requirements. The authors remain fully responsible for the accuracy, validity, and integrity of the content presented.

%
%
%
%

\end{document}